\pdfoutput=1

\documentclass[11pt]{article}

\usepackage[preprint]{acl}

\usepackage{times}
\usepackage{latexsym}

\usepackage[T1]{fontenc}

\usepackage[utf8]{inputenc}

\usepackage{microtype}

\usepackage{inconsolata}

\usepackage{graphicx}

\usepackage{booktabs}
\usepackage{enumitem}
\usepackage{subcaption}
\usepackage{xcolor}
\newcommand{\update}[1]{#1}

\title{Rethinking Evaluation Metrics for Grammatical Error Correction: \\Why Use a Different Evaluation Process than Human?}

\author{Takumi Goto, \
  Yusuke Sakai, \
  Taro Watanabe \\
  Nara Institute of Science and Technology \\
  \texttt{\{goto.takumi.gv7, sakai.yusuke.sr9, taro\}@is.naist.jp}}

\begin{document}
\maketitle
\begin{abstract}
One of the goals of automatic evaluation metrics in grammatical error correction (GEC) is to rank GEC systems such that it matches human preferences.
However, current automatic evaluations are based on procedures that diverge from human evaluation.
Specifically, human evaluation derives rankings by aggregating sentence-level relative evaluation results, e.g., pairwise comparisons, using a rating algorithm, whereas automatic evaluation averages sentence-level absolute scores to obtain corpus-level scores, which are then sorted to determine rankings.
In this study, we propose an aggregation method for existing automatic evaluation metrics which aligns with human evaluation methods to bridge this gap.
We conducted experiments using various metrics, including edit-based metrics, $n$-gram based metrics, and sentence-level metrics, and show that resolving the gap improves results for the most of metrics on the SEEDA benchmark.
We also found that even BERT-based metrics sometimes outperform the metrics of GPT-4.
\update{The proposed ranking method is integrated \textsc{gec-metrics}\footnote{A library for GEC evaluation proposed by ~\citet{goto2025gecmetricsunifiedlibrarygrammatical},~\url{https://github.com/gotutiyan/gec-metrics}.}}.
\end{abstract}

\section{Introduction}

\begin{figure}[t]
\centering
\includegraphics[width=0.48\textwidth]{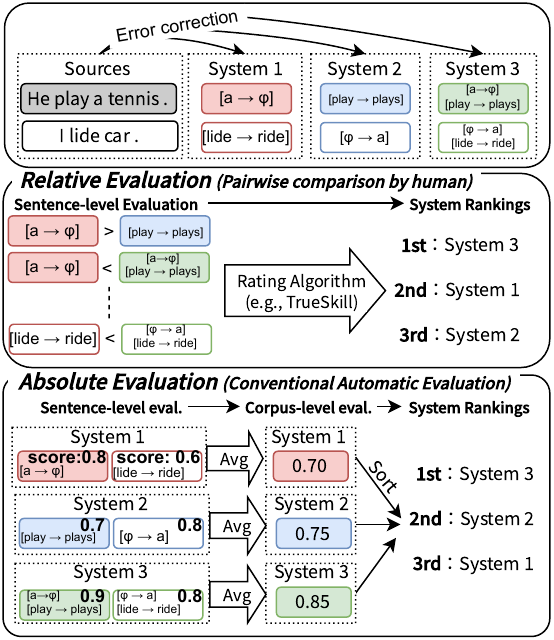}
\caption{An overview of current human and automatic evaluation when ranking three GEC systems based on a dataset containing two sentences. Each system output represents edits for simplicity.}
\label{fig:overview}
\end{figure}

Grammatical error correction (GEC) task aims to automatically correct grammatical errors and surface errors such as spelling and orthographic errors in text.
Various GEC systems have been proposed based on sequence-to-sequence models~\cite{katsumata-komachi-2020-stronger, rothe-etal-2021-simple}, sequence tagging~\cite{awasthi-etal-2019-parallel, omelianchuk-etal-2020-gector}, and language models~\cite{kaneko-okazaki-2023-reducing, loem-etal-2023-exploring}, and it is crucial to rank those systems based on automatic evaluation metrics to select the best system matching user's demands. 
Automatic evaluation is expected to rank GEC systems aligning with human preference, as evidenced by meta-evaluations of automatic metrics that assess their agreement with human evaluation~\cite{grundkiewicz-etal-2015-human, kobayashi-etal-2024-revisiting}.
For example, one can compute Spearman's rank correlation coefficient between the rankings produced by automatic and human evaluation, considering a metric with a higher correlation as a better metric.  

However, despite the clear goal of reproducing human evaluation, current automatic evaluation is based on procedures that diverge from human evaluation.  
Figure~\ref{fig:overview} illustrates the evaluation procedure for ranking three GEC systems using a dataset comprising two sentences.  
In human evaluation, corrected sentences generated for the same input sentence are compared relatively across system outputs, i.e., pairwise comparison, and the results are aggregated as rankings using rating algorithms such as TrueSkill~\cite{NIPS2006_f44ee263}.
In contrast, automatic evaluation estimates sentence-wise scores, then averages them at the corpus level and determines rankings by sorting these averaged scores.
As such, current automatic evaluation follows a procedure that deviates from human evaluation, contradicting the goal of reproducing human judgment. 
Intuitively, it would be desirable for automatic evaluation to follow the same procedure as human evaluation.

In this study, we hypothesize that resolving this gap will more closely align automatic evaluation with human evaluation.
Based on this hypothesis, we propose computing rankings in automatic evaluation using the same procedure as human evaluation, e.g., using TrueSkill after deriving pairwise estimates based on sentence-wise scores when human evaluation employs TrueSkill.
In our experiments, we conducted a meta-evaluation on various existing automatic evaluation metrics using the SEEDA dataset~\cite{kobayashi-etal-2024-revisiting}, that is a representative meta-evaluation benchmark. 
The results show that bridging the identified gap improves ranking capability for many metrics and that BERT-based~\cite{devlin-etal-2019-bert} automatic evaluation metrics can even outperform large language models (LLMs), GPT-4~\cite{openai2024gpt4technicalreport}, in evaluation.  
Furthermore, we discuss the use and development of automatic evaluation metrics in the future, emphasizing that sentence-level relative evaluation is particularly important for developing new evaluation metrics.

\section{Gap Between Human and Automatic Evaluation}

\subsection{Background}

Human evaluation has been conducted by \citet{grundkiewicz-etal-2015-human}, who manually evaluated systems submitted to the CoNLL-2014 shared task~\cite{ng-etal-2014-conll}, and by \citet{kobayashi-etal-2024-revisiting}, who included state-of-the-art GEC systems such as LLMs in their dataset.
In both studies, system rankings were derived by applying a rating algorithm to sentence-level pairwise comparisons.
Commonly used rating algorithms include Expected Wins~\cite{bojar-etal-2013-findings} and TrueSkill~\cite{NIPS2006_f44ee263, sakaguchi-etal-2014-efficient}. \citet{grundkiewicz-etal-2015-human} adopted Expected Wins as their final ranking method, whereas \citet{kobayashi-etal-2024-revisiting} used TrueSkill to determine the final ranking.
\citet{kobayashi-etal-2024-revisiting} also pointed out the importance of aligning the granularity of evaluation between automatic evaluation and human evaluation, but did not mention the procedure for converting sentence-level evaluation into system rankings.

Automatic evaluation is conducted using various evaluation metrics, including reference-based and reference-free approaches, as well as sentence-level and edit-based metrics.
Most of these metrics follow a procedure in which each sentence is assigned an absolute score, which is then aggregated into a corpus-level evaluation score.  
For example, sentence-level metrics such as SOME~\cite{yoshimura-etal-2020-reference} and IMPARA~\cite{maeda-etal-2022-impara} aggregate scores by averaging, while edit-based metrics such as ERRANT~\cite{felice-etal-2016-automatic, bryant-etal-2017-automatic} and GoToScorer~\cite{gotou-etal-2020-taking}, as well as $n$-gram-based metrics such as GLEU~\cite{napoles-etal-2015-ground, napoles2016gleutuning} and GREEN~\cite{koyama-etal-2024-n-gram}, aggregate scores by accumulating the number of edits or $n$-grams.
The corpus-level scores obtained through these methods can be converted into system rankings by sorting.

\subsection{How to Resolve the Gap?}

The gap can be simply addressed by using automatic evaluation metrics in the same manner as human evaluation.
Given that the SEEDA dataset uses TrueSkill as the aggregation method, we will close the gap by using TrueSkill for automated evaluation as well.
First, since existing automatic evaluation metrics compute sentence-wise scores, we convert these scores into pairwise comparison results.
For example, in the case illustrated in Figure~\ref{fig:overview}, the evaluation scores of 0.8, 0.7, and 0.9 for corrected sentences corresponding to the first sentence (``He play a tennis'') can be compared to produce pairwise comparison results similar to those in human evaluation.
Next, we compute system rankings by applying TrueSkill to the transformed pairwise comparison results.
In this study, we consider all combinations of pairwise comparisons for system set.
That is, given $N$ systems, a total of $N(N-1)$ comparisons are performed per sentence, and system rankings are computed based on these results including ties.

A similar method was employed by \citet{kobayashi-etal-2024-large}, but they did not mention the gap.
Also, their experiments used the TrueSkill aggregation for their proposed LLM-based metrics, but used conventional aggregation methods, e.g., averaging, for other metrics.
We discuss and organize the gap between human and automatic evaluation in detail, and then solve the gap by applying TrueSkill to all metrics for fair comparison.

\update{
Note that our method is explained using TrueSkill, which is used as the human evaluation method for SEEDA. If another meta-evaluation dataset uses a different aggregation method, such as Expected Wins, we should use that instead. We emphasize that our claim is the importance of aligning the aggregation methods of human and automatic evaluations, as even this simple practice has been largely overlooked so far.
}

\section{Experiments}  
\subsection{Automatic Evaluation Metrics}
We provide more detailed experimental settings for each metric in Appendix~\ref{sec:metric-config}.

\paragraph{Edit-based metrics}
We use ERRANT~\cite{felice-etal-2016-automatic, bryant-etal-2017-automatic} and PT-ERRANT~\cite{gong-etal-2022-revisiting}.
Both are reference-based evaluation metrics that assess at the edit level. 
When multiple references are available, the reference that yields the highest $F_{0.5}$ score is selected for each sentence.

\paragraph{$n$-gram based metrics}
We use GLEU+~\cite{napoles-etal-2015-ground, napoles2016gleutuning} and GREEN~\cite{koyama-etal-2024-n-gram}.
The $n$-gram overlap is checked among the input sentence, hypothesis sentence, and reference sentence.
When multiple references are available, GLEU+ uses the average score across all references, and GREEN uses the reference that yields the highest score is selected for each sentence.

\paragraph{Sentence-level metrics}
SOME~\cite{yoshimura-etal-2020-reference}, IMPARA~\cite{maeda-etal-2022-impara}, and Scribendi Score~\cite{islam-magnani-2021-end} are used.
All of them are based on small neural models such as $\textrm{BERT}_{\textrm{base}}$~\cite{devlin-etal-2019-bert} and designed as a reference-free metric that considers the correction quality estimation score as well as the meaning preservation score between the input and corrected sentences.

\begin{table*}[t]
    \centering
    \small
    \begin{tabular}{l|cc|cc|cc|cc}
    \toprule
        & \multicolumn{4}{c}{SEEDA-S} & \multicolumn{4}{c}{SEEDA-E}   \\
        & \multicolumn{2}{c}{\texttt{Base}} & \multicolumn{2}{c}{\texttt{+Fluency}} & \multicolumn{2}{c}{\texttt{Base}} &
        \multicolumn{2}{c}{\texttt{+Fluency}} \\
        Metrics & $r$ (Pearson) & $\rho$ (Spearman) & $r$ & $\rho$ & $r$ & $\rho$ & $r$ & $\rho$ \\
        \midrule
        \multicolumn{9}{l}{\textit{w/o TrueSkill}} \\
        ERRANT & 0.545 & 0.343 & -0.591 & -0.156 & 0.689 & 0.643 & -0.507 & 0.033 \\
        PTERRANT & 0.700 & 0.629 & -0.546 & 0.077 & 0.788 & 0.874 & -0.470 & 0.231 \\
        GLEU+ & 0.886 & 0.902 & 0.155 & 0.543 & 0.912 & 0.944 & 0.232 & 0.569 \\
        GREEN & 0.925 & 0.881 & 0.185 & 0.569 & 0.932 & 0.965 & 0.252 & 0.618 \\
        SOME & 0.892 & 0.867 & 0.931 & 0.916 & 0.901 & 0.951 & 0.943 & 0.969 \\
        IMPARA & 0.916 & 0.902 & 0.887 & 0.938 & 0.902 & 0.965 & 0.900 & 0.978 \\
        Scribendi & 0.620 & 0.636 & 0.604 & 0.714 & 0.825 & 0.839 & 0.715 & 0.842 \\
        \midrule
        \multicolumn{9}{l}{\textit{w/ TrueSkill}} \\
        ERRANT &\underline{ 0.763 }&\underline{ 0.706 }&\underline{ -0.463 }&\underline{ 0.095 }&\underline{ 0.881 }&\underline{ 0.895 }&\underline{ -0.374 }&\underline{ 0.231}\\
        PTERRANT & \underline{ 0.870 } & \underline{ 0.797 } & \underline{ -0.366 } & \underline{ 0.182 } & \underline{ 0.924 } & \underline{ 0.951 } & \underline{ -0.288 } & \underline{ 0.279 } \\
        GLEU+ & 0.863 & 0.846 & 0.017 & 0.393 & 0.909 &\underline{ 0.965 }& 0.102 & 0.486\\
        GREEN & 0.855 & 0.846 & -0.214 & 0.327 & 0.912 & 0.965 & -0.135 & 0.420\\
        SOME &\underline{ 0.932 }&\underline{ 0.881 }&\underline{ 0.971 }&\underline{ 0.925 }& 0.893 & 0.944 &\underline{ 0.965 }& 0.965\\
        IMPARA &\underline{ 0.939 }&\underline{ \textbf{0.923} }&\underline{ \textbf{0.975} }&\underline{ \textbf{0.952} }& 0.901 & 0.944 &\underline{ 0.969 }& 0.965\\
        Scribendi &\underline{ 0.674 }&\underline{ 0.762 }&\underline{ 0.745 }&\underline{ 0.859 }&\underline{ 0.837 }&\underline{ 0.888 }&\underline{ 0.826 }&\underline{ 0.912}\\
        \addlinespace
        GPT-4-E (fluency) & 0.844 & 0.860 & 0.793 & 0.908 & 0.905 & \textbf{0.986} & 0.848 & 0.987 \\
        GPT-4-S (fluency) & 0.913 & 0.874 & 0.952 & 0.916 & \textbf{0.974} & 0.979 & \textbf{0.981} & \textbf{0.982} \\
        GPT-4-S (meaning) & \textbf{0.958} & 0.881 & 0.952 & 0.925 & 0.911 & 0.960 & 0.976 & 0.974\\
        \bottomrule
    \end{tabular}
    \caption{Correlation with human evaluation using the SEEDA dataset. \textit{w/o TrueSkill} refers to the conventional evaluation procedure, while \textit{w/ TrueSkill} represents the proposed evaluation procedure. Improvements over the conventional procedure are \underline{underlined}, and the highest value in each column is highlighted in \textbf{bold}. The GPT-4 results refer to those reported in \citet{kobayashi-etal-2024-revisiting}.}
    \label{tab:results}
\end{table*}

\subsection{Meta-Evaluation Method}  
We use the SEEDA dataset~\cite{kobayashi-etal-2024-revisiting} for meta-evaluation.
Meta-evaluation results are reported based on human evaluation results using TrueSkill for both the sentence-level human-evaluation, SEEDA-S, and the edit-level human-evaluation, SEEDA-E.
Additionally, we also report results for both the \texttt{Base} configuration, which excludes the fluent reference and GPT-3.5 outputs that allow for larger rewrites, and the \texttt{+Fluency} configuration, which includes them.  

Furthermore, we evaluate the robustness of the calculated rankings using window analysis~\cite{kobayashi-etal-2024-large}.
The window analysis computes correlation coefficients only for consecutive $N$ systems, after sorting systems based on human evaluation results. 
This allows us to analyze whether automatic evaluation can correctly assess a set of systems that appear to have similar performance from the human evaluation.
In this study, we perform it with $N=8$ for 14 systems corresponding to the \texttt{+Fluency} configuration, and report both Pearson and Spearman correlation coefficients.  
That is, correlation coefficients are computed for the rankings 1 to 8, 2 to 9, $\dots$, and 7 to 14 from human evaluation.

\subsection{Experimental Results}  
Table~\ref{tab:results} shows the results of the meta-evaluation.
The upper group presents evaluation results based on the conventional method of averaging or summing, and the bottom group presents results evaluated using TrueSkill, which follows the same evaluation method as human evaluation.
The bottom group includes the evaluation results based on GPT-4 reported by \citet{kobayashi-etal-2024-large}, which correspond to the state-of-the-art metrics.  

The overall trend indicates that using TrueSkill-based evaluation improves the correlation coefficients for most of metrics. 
In particular, the results of IMPARA in the SEEDA-S and \texttt{+Fluency} setting outperformed those of GPT-4 results.  
Additionally, ERRANT showed an improvement of more than 0.2 points in many configurations.  
These results show that using automatic evaluation metrics with the same evaluation procedure as human evaluation makes the ranking closer to human evaluation.
In other words, the existing automatic evaluation metrics were underestimated in the prior reports due to the gap in the meta-evaluation procedure.

\update{
Unlike edit-level or sentence-level metrics, no effect was observed in $n$-gram-level metrics such as GLEU+ and GREEN.
This stems from the poor sentence-level evaluation performance of $n$-gram based metrics.
The BLEU paper~\cite{papineni-etal-2002-bleu}, which is a $n$-gram-level metric for machine translation and the basis of the GLEU+, notes that the brevity penalty can excessively penalize scores for short sentences, and uses corpus-level lengths to address this issue.
Since GLEU+ also employs a brevity penalty, it cannot accurately calculate sentence-level scores depending on sentence length. This is a serious issue for TrueSkill-based aggregation because the quality of sentence-level scores directly affects the quality of system rankings.
Furthermore, while GREEN does not use a brevity penalty, the score can become unstable as the ``$n$'' for $n$-gram increases, especially for short sentences. This has a negative impact on the geometric mean among $n$-gram scores, which is the final score.
An ideal metric should provide evaluation results that better align with human judgments when ranking systems in the same way humans do. Given this premise, our results suggest a human alignment issue of $n$-gram-level metrics.
}

\begin{figure}[!t]
\centering
  \begin{minipage}[b]{0.48\textwidth}
    \centering
    \includegraphics[width=0.99\textwidth]{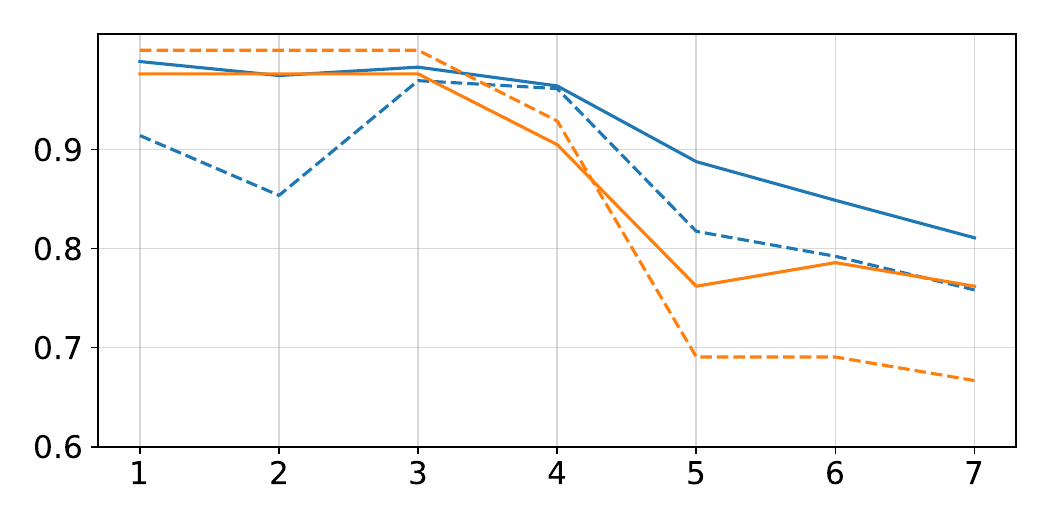}
    \subcaption{IMPARA}
    \label{fig:window-impara}
  \end{minipage}
  \begin{minipage}[b]{0.48\textwidth}
    \centering
    \includegraphics[width=0.99\textwidth]{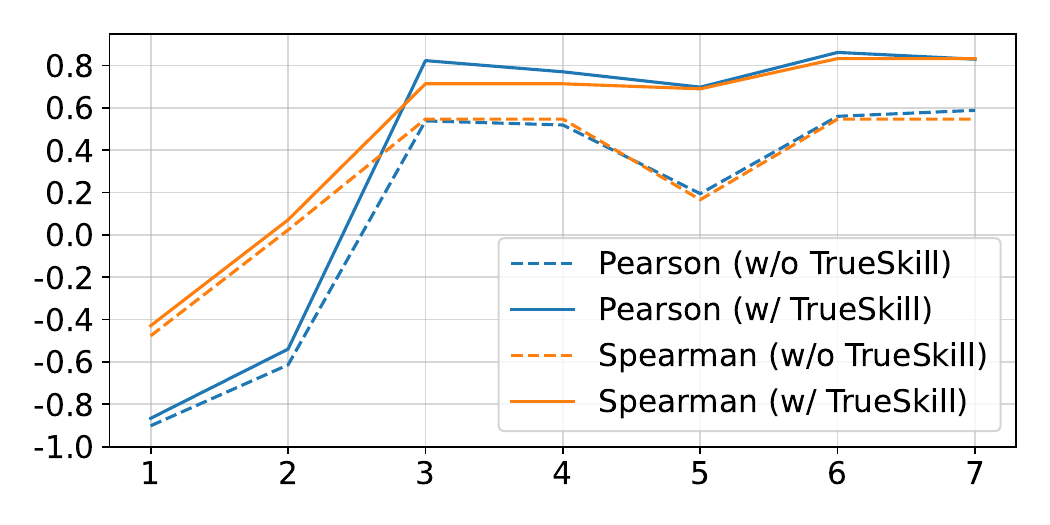}
    \subcaption{ERRANT}
    \label{fig:window-errant}
  \end{minipage}
  \caption{The results of the window analysis for $N=8$ are shown. The x-axis represents the starting rank of human evaluation. For example, $x=2$ shows the results for the systems ranked 2nd to 10th in human evaluation.}
    \label{fig:window-analysis}
\end{figure}

Figure~\ref{fig:window-analysis} shows the results of the window analysis for IMPARA and ERRANT measured on SEEDA-S and SEEDA-E, respectively.
From Figure~\ref{fig:window-impara}, it can be seen that IMPARA particularly aligns with human evaluation in the lower ranks.  
The Pearson correlation coefficient also showed an improvement in the evaluation results for the top systems as well.
Since the top systems include GEC systems that are largely rewritten, such as GPT-3.5, this characteristic is useful, considering that LLM-based correction methods will become popular in the future.
Figure~\ref{fig:window-errant} shows that ERRANT consistently showed improved correlation coefficients with the proposed method, but still struggled with evaluating the top systems.
For edit-based evaluation metrics, it is still considered difficult to assess such GEC systems even with the evaluation method aligned with human evaluation\footnote{Using a larger number of references may solve this issue.}.

\section{Conclusion}

In this study, we focused on the fact that human evaluation aggregates sentence-level scores into system rankings based on TrueSkill, while automatic evaluation uses a different evaluation, and we proposed to use TrueSkill in automatic evaluation as well.
Experimental results with various existing metrics showed improvements in correlations with human evaluation for many of the metrics, indicating that agreement on the aggregation method is important.
\update{
Our core statement in this paper is not about using TrueSkill, but rather the importance of using the same aggregation method as human evaluation. 
For instance, if future meta-evaluation datasets switch the aggregation method for human evaluation to averaging sentence-level scores, then automatic evaluation should likewise adopt the same approach.
}

Given the discussion so far,
\update{
at least in the current situation of GEC evaluation,
} we recommend transitioning the aggregation method from averaging or summing to using a rating algorithm, such as TrueSkill.
We also recommend that evaluation metrics should be developed that allow for accurate sentence-wise comparisons.
This is evidenced by the fact that IMAPARA achieves a higher correlation coefficient than SOME in Table~\ref{tab:results}.
In fact, IMAPARA is trained to assess the pairwise comparison results, whereas SOME is trained to evaluate sentences absolutely.

\section*{Limitations}
\paragraph{Use for Purposes Other Than System Ranking}  
The proposed method is designed for system ranking and cannot be used for other types of evaluation, such as analyzing the strengths and weaknesses of a specific system. For instance, when analyzing whether a model excels in precision or recall, it is more useful to accumulate the number of edits at the corpus level, as done in existing evaluation methods.

\paragraph{Reproducing the Outputs of Compared GEC Systems}  
Since the proposed ranking method requires inputting all GEC outputs being compared, it is necessary to reproduce their models.
This point is different from existing absolute evaluation methods, where previously reported scores can be cited.
While this may seem burdensome for researchers, it can also be seen as an important step toward promoting the publication of reproducible research results.

\section*{Ethical Considerations}
When the metric contains social biases, the proposed method cannot eliminate that bias and may reflect that bias in the rankings.
However, we argue that this problem should be resolved as a metric problem.

\section*{Acknowledgments}
\update{
We sincerely thank the anonymous reviewers
for their insightful comments and suggestions.
This work has been supported by JST SPRING. Grant Number JPMJSP2140.
}

\bibliography{custom}

\appendix

\section{Detailed Experimental Settings for Evaluation Metrics}\label{sec:metric-config}

We used \textsc{gec-metrics}~\cite{goto2025gecmetricsunifiedlibrarygrammatical} for the implementation. 
The detailed experimental settings are as follows:

\begin{description}[leftmargin=1em, itemsep=0em, topsep=0.5em]
    \item[ERRANT] Evaluations were conducted using the Python module \texttt{errant==3.0.0} with the span-based correction setting. Although the edits in the CoNLL-2014 references were manually annotated, they were re-extracted using ERRANT. This is important for an evaluation based on consistent edits.
    \item[PT-ERRANT] We employed the $F1$ score of BERTScore for the weighting. The baseline rescaling was adopted, and IDF adjustment was not performed. We used \texttt{bert-base-uncased} for the BERT model. These settings are consistent with those of the official implementation~\footnote{\url{https://github.com/pygongnlp/PT-M2}}. Similar to ERRANT, we re-extracted reference edits via \texttt{errant} module.
    \item[GLEU+] We used \textbf{GLEU+} with n-grams up to 4-grams and 500 iterations during reference sampling. 
    \item[GREEN] Similarly, we used \textbf{GREEN} with n-grams up to 4-grams, utilizing the $F_{2.0}$ score, following those employed by \citet{koyama-etal-2024-n-gram}.
    \item[SOME] We used the official models for grammaticality, fluency, and meaning preservation, with respective weights of 0.55, 0.43, and 0.02. These weights correspond to those tuned by \citet{yoshimura-etal-2020-reference} for sentence-level evaluation performance.
    \item[IMAPARA] As a pre-trained quality estimation model was not publicly available, we newly constructed it through reimplementation and experimentation. Following \citet{maeda-etal-2022-impara}, the CoNLL-2013 dataset~\cite{ng-etal-2013-conll} was used as a seed corpus and was split into training, development, and evaluation sets with an 8:1:1 ratio. We fine-tuned \texttt{bert-base-cased}, and followed \texttt{BertForSequenceClassification} from the Transformers library~\footnote{\url{https://github.com/huggingface/transformers}}~\cite{wolf-etal-2020-transformers} for the classifier architecture. This corresponds to transforming the \texttt{CLS} representation of the BERT model into a real value via a single projection layer. During inference, \texttt{bert-base-cased} was used for the similarity estimation model, with a threshold set to 0.9.
    \item[Scribendi Score] GPT-2~\cite{radford2019language}\footnote{\url{https://huggingface.co/openai-community/gpt2}} was utilized as the language model, and the threshold for the maximum of the Levenshtein-distance ratio and token sort ratio was set to 0.8.
\end{description}

\end{document}